\documentclass[11pt,a4paper]{article}
\usepackage[hyperref]{acl2019}
\usepackage{times}
\usepackage{latexsym}
\usepackage{graphicx}

\usepackage{url}

\aclfinalcopy % Uncomment this line for the final submission
\title{Modeling Confidence in Sequence-to-Sequence Models}

\author{Jan Niehues \\
  Department of Data Science \\ and Knowledge Engineering (DKE) \\
  Maastricht University\\
  \texttt{jan.niehues@maastrichtuniversity.nl} \\\And
  Ngoc-Quan Pham  \\
  Institute of Anthropomatics \\
  Karlsruhe Initute of Technology \\
  \texttt{ngoc.pham@kit.edu}}

\date{}

\begin{document}
\maketitle
\begin{abstract}

Recently, significant improvements have been achieved in various natural language processing tasks using neural sequence-to-sequence models. 
While aiming for the best generation quality is important, ultimately it is also necessary to develop models that can assess the quality of their output. 

In this work, we propose to use the similarity between training and test conditions as a measure for models' confidence. 
We investigate methods solely using the similarity as well as methods combining it with the posterior probability. 
While traditionally only target tokens are annotated with confidence measures, we also investigate methods to annotate source tokens with confidence. 
By learning an internal alignment model, we can significantly improve confidence projection over using state-of-the-art external alignment tools.
We evaluate the proposed methods on downstream confidence estimation for machine translation (MT). We show improvements on segment-level confidence estimation as well as on confidence estimation for source tokens.
In addition, we show that the same methods can also be applied to other tasks using sequence-to-sequence models. 
On the automatic speech recognition (ASR) task, we are able to find 60\% of the errors by looking at 20\% of the data.
% \liz{I don't know that this sentence makes sense to me} 

\end{abstract}

\section{Introduction}

Deep learning methods have significantly increased the quality of natural language generation tasks such as Machine Translation~(MT). However, when deployed in a production environment, understanding the model's confidence and how well it correlates with output quality is as important as training the best models. 

While humans are often capable of estimating whether their decisions are sensible or produced by random guesses, it is often not possible to know how confident deep learning models are with respect to their output ~\cite{Gal2016Uncertainty}. However, information regarding confidence can be essential in production scenarios. In cases with a human-in-the-loop, confidence can be used to identify the parts of the machine output that require human intervention, e.g. in post-editing for machine translation or to guide reformulation of the original input to simplify the task for sequence-to-sequence models. 

Intuitively, models should have higher confidence towards data points that are similar to their training data. 
Motivated by this, our first contribution is an autoencoder network that is applied as an extension to the sequence-to-sequence models to measure the training-testing discrepancy. 
In contrast to methods that directly compare the test and training data to generate confidence scores, we do not need to store the whole training data, thereby enabling our method to scale to larger datasets and tasks. 

Motivated by the successful application of posterior probabilities for confidence estimation in statistical machine translation (SMT) \cite{ueffing2007} and traditional ASR systems \cite{Siu1999}, our second contribution is a combination our approach with this prior approach.

Traditionally, confidence estimation has been defined as a task of assessing the quality of the whole sequence of words in the target sentence. 
Especially when evaluating translations, there are also several cases when it can be  very beneficial to estimate how well the source words are translated beyond coverage. 
For example, a person only speaking the source language might be able to reformulate the source sentence, if he knows that the system has difficulties with certain words. 
As our third contribution, we present a method to estimate the alignment between source and target tokens in complex sequence-to-sequence models. 
We can show that this strongly outperforms external state-of-the-art alignment methods.

Our experiments shows that in machine translation, the posterior probabilities can be competitive with automatic metrics in terms of correlation with human evaluation. 
For speech recognition, we are able to find 60\% of the errors by looking at 20\% of the data. 

% \liz{I don't get this sentence, rephrase or make more explicit? learning from 20\% of the data you can find the rest? how would this work across vocab?}
%%% Quan: tomorrow morning I will try to get this sentence. 

%This paper is structured in the following way: In the next section we start with an overview of the framework we used for confidence estimation. In Section \ref{traniningSimilarity}, we introduce the methods to use the training similarity for confidence estimation. In the following, we describe methods to map target estimations to source estimations. Section \ref{sec:eval} will give a details description on how we evaluated the confidence measures. Next, we will describe the experiments. We will end with related work and a conclusion

\section{Confidence Estimation Task} 

Depending on the use case, there are different ways to define the task of confidence estimation. Furthermore, there is no clear separation between confidence estimation and quality estimation. 
% \liz{not sure how i feel about this sentence. i agree the line between them isn't always clear, but a model can be confidently wrong; they are overlapping but distinct ?}
A first important dimension is the granularity of the predictions. We investigate three different use cases in this work, described in the next three subsections in greater detail.

Previous methods differ in whether they predicting continuous values or discrete labels. In this work, we will predict continuous values, but evaluate against gold standard  labels. In Section \ref{sec:eval}, we describe in detail how we map continuous predictions to discrete labels.

In addition, previous methods differ in whether they can be trained on gold standard labels or if no annotated training data is available. 
Training data is a particular challenge in confidence estimation since annotations are associated with the output of a particular model. 
Therefore, this raises the question whether the task is to estimate the quality of \textit{any} model, or of a particular one. 
This has implications for whether we can use model internal information or not. 
In this work, we focus on the situation where we want to estimate the confidence of a particular model, using internal information. 
Since in a realistic real-world scenario we are not able to collect annotated data for each model we are interested in, we further do not use any labeled training data.

\subsection{Granularity}

First, the confidence of the whole output sequence can be estimated. Given an input sequence $X=x_1 \ldots x_{I_w}$ and an output sequence $Y=y_1 \ldots y_{J_w}$, the model estimates quality $c$ for the whole sequence.
%, which can be either  a binary value expressing whether errors exist in the sequence or a floating point number describing the quality of the translation.
We will present several methods that calculate a sequence of confidence estimations $c'_1,...,c'_L$. Therefore, we need an additional aggregation function for the sequence confidence estimates. In all our experiments, we are using the minimum as the aggregation function.

%\subsection{Output token confidence}

In some use cases, it is important to get more fine-grained quality estimation.
To be specific, we aim at estimating the confidence of every target token $x_j$ instead of one single score for the sequence.  Given an input sequence $X=x_1 \ldots x_{I_w}$ and an output sequence $Y=y_1 \ldots y_{J_w}$, the output will be a sequence of quality estimations $C=c_1 \ldots c_{J_w}$. One additional challenge is that we might be interested in the confidence using a different granularity than the predicted by the model $c'_1,...,c'_L$ (with $L \ne J_w$). 
For example, the user is interested in word-based confidence, while the system uses subword units. 
% \liz{this is the opposite of the first sentence in the paragraph. Quan: less than sequence but different level of granularity.}
In this case, we assume to have a mapping $m$ between the positions $1 \ldots J_w$ and $ 1 \ldots L$. In the example of subwords, this is straightforward because segmentation is recoverable. Then, we also need an additional aggregation function for the confidence estimates. We estimate the confidence $c_{j}$ by $agg_{m(l)=j}(c_{l})$. For this type of aggregation we also use the minimum.

%\subsection{Input token confidence}

In machine translation, it is not only the confidence at the output level that is of interest, but also how adequately each individual source token is translated. 
From an application point of view, when the machine translation is used in an interactive scenario, this feature for example enables the user to reformulate the source sentence in order to avoid phrases that the system is not able to handle. 

Formally, given an input sequence $X=x_1 \ldots x_{I_w}$ and an output sequence $Y=y_1 \ldots y_{J_w}$, the model estimates a sequence of confidence measures $C=c_1 \ldots c_{I_w}$. Therefore, in this case, given the estimation of the model  $c'_1,...,c'_L$, we need a mapping $m$ between the positions $1 \ldots I_w$ and $ 1 \ldots L$.

\subsection{Posterior Probabilities}

As a baseline for our experiments, we use the posterior probabilities. The intuition behind this technique is that the model will distribute the probability mass over several outputs in low-confidence situations. In contrast, if the model is confident about its prediction, it should assign a high probability to the prediction. 

%One main advantage of using sequence-to-sequence models instead of SMT is that this model can assign a probability to every possible output. We can therefore use this model for estimating its own quality as well as for estimating the quality of the output of a different system.

Formally, given an input sequence $X=x_1 \ldots x_{I_w}$ and an output sequence $Y=y_1 \ldots y_{J_w}$, we first define the input tokens $X'=x'_1 \ldots x'_{I_y}$ and an output sequence $Y'=y'_1 \ldots y'_{J_t}$ (e.g. by using subwords). The encoder will first calculate a sequence of hidden states $E=e_1, \ldots e_{I_t} = ENC(X')$. Secondly, we predict the target hidden states $D=d_1, \ldots d_{J_t} = DEC(E,Y')$. Finally, we can use the posterior probabilities $P=p_1, \ldots, p_{J_t}$ calculated by:

\begin{equation}
\label{pp}
    p_i = softmax(FF(d_i))[y'_i]
\end{equation}
where $FF$ is a linear transformation and $[k]$ indicates the $k$-th element of the vector. By using $P$ for $C'$ as described in Section 2.1, we can calculate now a sequence confidence or an output confidence.

\section{Training similarity}
\label{traniningSimilarity}

The similarity between test input and the examples seen in training is an important indication for the model's performance. Intuitively, models should be better at predicting examples similar to their training data than examples very different from the training data.

\subsection{Approaches to measure similarity}

Two sentences can be similar in many ways. Therefore, there are also many ways to estimate the similarity between sentences. For our use case, it is important how similar the sentence representation generated by the translation systems is. Hence, we use the internal representations of the neural machine translation model to measure the similarity of the sentences.

In an NMT system, there are different representation levels which can be used to measure the similarity of the sentence. For example, we can use the final encoder hidden states, the final decoder hidden states, or the context vectors. As motivated in the introduction, one interesting use case for using confidence is to find difficult source segments, so that the user can rewrite them.  For this case, we concentrate on the encoder hidden states. 
%By aligning each hidden state to a source word we identify the difficult parts in the source sentence.

We measure the training-test similarity as follows:
First, we run the encoder on the source side of the training data and store the encoder hidden representation (top layer) for every sentence $k$ ($E^k=e^k_1, \ldots, e^k_{I^k_t}$). 
Second, we calculate the hidden representations of the test sentences ($E^{tst}=e^{tst}_1, \ldots, e^{tst}_{I_T^{tst}}$) and used approximate k-nearest neighbor search (implemented in the Annoy\footnote{https://github.com/spotify/annoy} toolkit). 
%Then the $L^2$ distance is measured between the test representation and the nearest neighbor to express the difficulty. 

We investigated two methods to estimate the similarity, one on the sentence level and one on token level.
First, we use the distance to the overall most similar training sentence by using the average vector of the encoder hidden states for the training as well as for the test data. Formally:
\begin{equation}
s=\min\limits_{k \in train} L^2(avg(e^k_1, \ldots, e^k_{I_k}),avg(e^{tst}_1, \ldots, e^{tst}_{I_{tst}}))
\end{equation}
Then we use $s$ directly as the sequence confidence $c$ from Section 2.1.

The second method is to estimate the confidence for each source token. This is achieved by finding the nearest neighbor for each hidden encoder state $e^{tst}_i$. 
\begin{equation}
s_i\min\limits_{k \in train;i_k \in {1, \ldots, I_k}} L^2(e^k_{i_k},e^{tst}_i)
\end{equation}
By using $S=s_1 \ldots s_{I^{tst}}$ as $C'$ in Section 2.1, we can calculate a sequence confidence or an confidence for each input token.

% \subsubsection{Approximating similarity}
 \subsection{Similarity estimation}
% \liz{can you reference results/numbers to show that below here?}
% While the aforementioned approach is able to calculate the difference between the test and training sentence accurately, 
The main disadvantage of aforementioned method is that we need to calculate and store the hidden representation of all training examples. 
Such storage consumption is non-trivial even for small datasets like the TED corpus and it is infeasible for large-scale sequence-to-sequence models.

Therefore, we also investigate methods to approximate the distance without storing the hidden states for the whole training data. Here we propose to approximate this distance by using autoencoders.
%which is firstly established by training an autoencoder on a particular hidden representation of the sequence to sequence model. 
The autoencoder will be able to reconstruct typical hidden states seen in the training data, while the reconstruction of unusual hidden states will be less exact.

\begin{figure}[ht!]
\centering
\includegraphics[width = 0.1\textwidth]{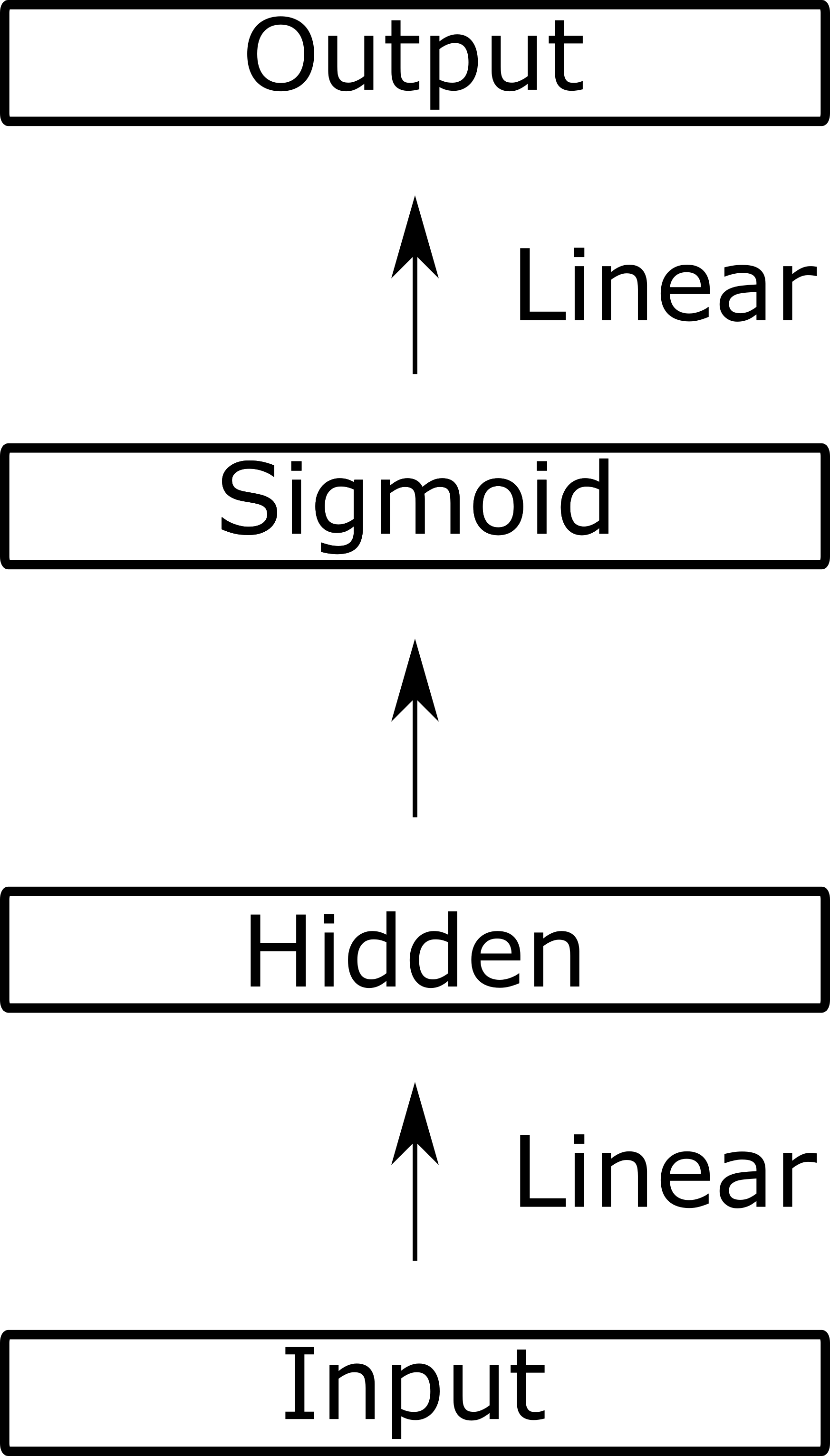}
\caption{\label{fig:autoencoder} Architecture of the autoencoder}
\end{figure}

As shown in Figure \ref{fig:autoencoder}, we are using an autoencoder with a single hidden layer. In our experiments, we investigate different hidden sizes of the autoencoder. Afterwards, we apply the sigmoid activation function before predicting the output.

Next, we then can use the quality of the reconstruction as a measure of the model's confidence in its predictions. We found that it is possible to get the confidence qualitatively by measuring the $L^2$-distance between the hidden representation and its reconstruction.
\begin{equation}
    s^e_i = L2(e_i,Auto(e_i))
\end{equation}
As for the direct measurement, we can use $S^e=s^e_1 \ldots s^e_{I^{tst}}$ as $C'$ to calculate the confidence of the sequence or for each input token. Furthermore, by using the decoder states $D$ instead of the encoder states $E$, we can calculate $S^d$ accordingly and use it to estimate the sequence or target token confidence.
%Compared to the direct measurement, we are able to approximate the difference now also for large data sets. 

\subsection{Combining both approaches}

While using similarity measurements is able to estimate the quality of the whole sequence as well as of parts of the sequence, it also has two drawbacks: First, in the $L^2$-norm all dimensions are equally important, while this might not be the case for the final prediction of the words. Second, we are only looking at the similarity between training and test condition, but ignoring that some outputs might be inherently more difficult to predict than others.

Therefore, we combine both techniques and thereby minimize their respective drawbacks. To do so, the hidden representation is first replaced by the reconstruction generated by the autoencoder. After that, we calculate the probabilities based on these reconstructed hidden representations. If we use the autoencoder on the decoder states, Equation \ref{pp} needs to be modified to:
\begin{equation}
\label{pp_sim}
    p^s_i = softmax(FF(Auto(d_i)))[y'_i]
\end{equation}
 By using $P^s$ for $C′$ as described in Section 2.1, we can calculate now a sequence confidence or an output confidence. Similarly, we can replace $Auto(d_i)$ by $d'_i$ with $D' = DEC(Auto(E),Y')$ to use the autoencoders on the encoder hidden states. It is important to note, that the similarity approximated by the encoder hidden states can be used for source token confidence estimation, while the combination of autoencoders on the source hidden state and posterior probabilities can only be used for target token confidence estimation. One advantage of this combination is that no additional parameters are introduced.

\section{Alignment}
\label{alignment}
While the previously presented models are all able to generate confidence measures for each target token, only the distance-based similarity measures are able to also generate scores for the source tokens. In order to generate source token confidence qualitatively, a straightforward approach is to use word alignment to map the confidence score from the target side to the source.

Our baseline for these experiments uses the IBM4 GIZA alignment model \cite{Och03} to map the posterior probabilities and the combined approach's confidence estimations from target to source tokens. If several target tokens align to the same source token, we again use the minimal confidence. 

% \liz{minimal confidence? GIZA score or model confidence score? shouldn't it be max either way?}

%\subsection{Model}

Motivated by our autoencoding approach to measure similarity between training and test data, we investigate similar approaches to model the alignment between source and target tokens. In this case, we used a model to predict a target hidden state $d_j$ given a source state $e_i$. If a source word aligns to a target word, it should be possible to predict this target word primarily based on this source word. Therefore, we choose the same architecture as for the autoencoder. We use the source hidden state to predict a target hidden state. Then, we compare the predicted hidden state to all decoder hidden states and describe the alignment strength between the source and target hidden state using the cosine similarity between the predicted hidden state and the target hidden state. Let $NN()$ be the neural network-based predictor. We then calculate the alignment by:
\begin{equation}
    a'_{ij} = \cos sim(NN(e_i),d_j)
\end{equation}

Based on the alignment scores, we created an alignment matrix by aligning each source word to the target word with the strongest link according to Equation~\ref{eq:2}.

\begin{equation}
    \label{eq:2}
    a(i) = \min\limits_{j \in {1, \ldots, J}} a'_{ij}
\end{equation}

%\subsection{Training}

Since there are not confidence labels with aligned source and target words available, we cannot simply train the neural network. Inspired by the GIZA model, we utilize the EM algorithm for training.
Given an alignment $a*$, we can train our model using the following MSE-based loss function:
\begin{equation}
    MSE(NN(e_i),d_{a*_i})
\end{equation}
This can be extended for soft alignments $a$ to:
\begin{equation}
    MSE(NN(e_i),\sum_j a_{ij}d_{j})
\end{equation}
This corresponds to the M-Step in the EM algorithm. To be able to train the model using this loss function, we need to estimate an alignment $a$ in the E-step.
Given the source representation $e_1,\ldots e_I$ of a sentence, we use the predictor to calculate the prediction $p_1,\ldots p_I$. Based on this, we calculate the alignment similarities $a'_{ij}$ based on cosine similarities between $p_i$ and the decoder hidden states $d_j$. In order to prevent the model from learning to collapse into aligning all source words to the most obvious words e.g. the period at the end of the sentence, we normalize them to probabilities for each target word. ($a'_{ij} = a_{ij}/ \sum_{i'=1}^I(a_{i'j})$).

\section{Evaluation}
\label{sec:eval}

In this work, we evaluate the ability of sequence-to-sequence models to estimate their confidence in their own output on two different tasks: MT and ASR.

It is necessary to define a gold standard for the evaluation. For ASR, there is only one ground truth. 
Accordingly, we can label each output word from the model as correct/substitution/deletion/insertion. 
Our confidence measurement is then done on he word level (predicting whether the word is correct or not)~\footnote{If a word in the reference was deleted, we marked the previous and next word also as an erroneous word.}.

For machine translation, a single correct translation for each source sentence does not exist. To account for this, our experiments are carried out in the following way: We collected annotations with incorrectly translated source words for 1177 sentence pairs, resulting in $39.93\%$ of the source sentences containing mistranslated words. We were not able to test our methods on existing quality estimation data sets, as we cannot access internal model information for this data.
% \liz{.. because you couldn't retrain on the data they had been trained on?}.

Given the reference labels, the next step is to measure the quality of the confidence measures. In our experiments, we use four different measures. The first possible scenario is that, we assume that the user has a fixed amount of time and wants to maximize the improvements. Therefore, we calculate the confidence score for all the test data and look at the 10\% and 20\% of the test data that the model has given the lowest confidences. Then, we measure what percentage of errors according to the reference are found in this part of the data. 
%Consequently, a system randomly selecting sentences would find 10\% and 20\% of the errors.

In another scenario, we want to dynamically correct as many sentences as would be beneficial. This can be measured using the F-Score. Since we need to map the confidence scores to labels, we have the additional challenge  of finding a good threshold for when to assign the label ``high confidence'' or ``low confidence'' to an output sentence/word. Therefore, we report oracle F-Scores using an optimal threshold found on the test data. Furthermore, we evaluate an approach to find this threshold in an unsupervised manner:
While our baseline system uses beam search with beam=$8$, we also perform greedy decoding. We assume that the model is not confident if the beam search leads to a different outcome from the greedy decoding, and create pseudo-labels where each segment or token is labeled wrong if the results of beam search and greedy search differ. Then, we select the optimal threshold based by comparing the predict scores and these pseudo-labels and evaluate the approach on the real labels.

\section{Experiments}   

%\subsection{Model configurations}

The sequence-to-sequence models in our work are based on the state-of-the-art Transformer architecture~\cite{vaswani2017attention}. We followed the model configuration with the learning rate schedule from the~\textit{Base} configuration in the original work. The number of layers is adapted for each task for the best performance possible and will be reported respectively. The autoencoders are implemented on top of the Transformer (with PyTorch~\cite{paszke2017automatic}) using one hidden layer with different sizes and sigmoid activation function. \footnote{https://github.com/isl-mt/NMTGMinor} 
The MT model is a 12-layer Transformer trained on the German-English TED corpus \cite{cettolo2012} with the development set and test set from the IWSLT 2017 evaluation campaign. The data is preprocessed with Moses tokenization, true-casing and segmented with byte-pair encoding \cite{Sennrich2016} with 40K codes. The model achieves a BLEU score of $28.82$ on the development set and $30.63$ on the test data.

We conducted further ASR experiments on the Switchboard-1 Release 2 (LDC97S62) corpus, which contains over 300 hours of speech. The Hub5'00 evaluation data (LDC2002S09) was used as our test set. On this set, we are especially interested in the influence of the model performance on the quality estimation. Therefore, we trained 4 different models with 4,8,12 and 24 layers. These models achieve a WER of $20.8$, $14.8$, $13.0$ and $12.1$ on the Switchboard test set respectively, and $33.2$, $25.5$, $23.9$ and $23.0$ on the Callhome set.

\subsection{Machine translation results}

\begin{table*}[ht]
 \begin{center}
   \begin{tabular}{|l|c|c|c|c|} \hline 
Model & 10\% & 20\% & Oracle & Pseudo-label \\ \hline
BEER &  10.43 & 24.68 & 61.24 & 61.10 \\
Prob &  \textbf{17.66} & 33.19 & 64.66 & 63.96  \\ \hline
Enc Sent Distance & 17.02 & 31.91 & 62.95 & 61.73\\ 
Enc Distance & 16.81 & 32.77 & 63.66 & 62.64 \\ \hline
Enc Auto 128 & 13.62 & 25.74 & 60.13 & 59.64\\
Enc Auto 256 & 14.47 & 29.57 & 61.94 & 61.37 \\
Enc Auto 512 & 13.62 & 25.74 & 60.50 & 59.61 \\ \hline
Dec Auto 128 & 15.11 & 30.00 & 62.87 & 61.63  \\
Dec Auto 256 & 16.17 & 31.91 & 63.96 & 61.43  \\
Dec Auto 512 &  15.11 & 31.28 & 62.92 & 60.31  \\ \hline
Enc Auto 256 + Prob &  16.38 & \textbf{34.04} & 64.48 & 64.15 \\
Dec Auto 256 + Prob & 17.02 & \textbf{34.04} & \textbf{65.92} & \textbf{65.54}  \\ \hline

\end{tabular}   
\end{center}
\caption{\label{table:smallMT:sent}Segment-level confidence estimation for MT. First two columns: Percentage of found errors when selecting 10\% and 20\% of the data; Final two columns: F-score when using oracle threshold and thresholds optimized on pseudo-labels}
\end{table*}

The first concern in the experiments is the performance on segment-level quality estimation for machine translation. 
%As described in Section \ref{sec:eval}, we used four different scores to describe the performance of the systems. 
The results are summarized in Table \ref{table:smallMT:sent}.

Two baseline systems are presented in this experiment. To measure the difficulty of the task, we use the BEER evaluation metric as comparison, which has been performing competitively in the WMT Metric evaluations~\cite{stanojevic2014beer}. It is important to note that the metric has access to the reference translation, while the confidence measure do not. Even with this advantage, the metric does not clearly outperform a random baseline, showing the difficulty of the task. Using the model's posterior probability, we can improve on all four types of confidence measure. Among the 10\% of the sentences with the lowest confidence, this method was able to find 17.66\% of the sentences with errors. For this task, this is further the best performance. This confirms our hypothesis that the posterior probabilities can be reliable for modelling the system's confidence.

Proceeding to experiments shown in the next two lines, we evaluate the ability of using the similarity between test and training data as a measure for confidence. Although not performing as well as the posterior probabilities, the data difference is a good estimator for the task difficulty and the confidence of the model. When comparing a single sentence representation (Enc Sent Distance) and the token representation (Enc Distance) in the next line, the second one outperforms the first one, except for the top 10\%. Therefore, it seems to be important to measure the distance of each individual token and not only of the whole sentence. 

Motivated by these results, we trained the autoencoders on the individual tokens and not on the whole sentence and used the autoencoder networks on the source hidden representation to estimate the performance. We analyze the influence of the size of the bottleneck of the autoencoder.  The network with bottleneck size of $256$ (Enc Auto 256), which is half the size of the input size, managed to get the best performance in all measures. While we see a drop in performance due to the approximation, e.g. from 32.77\% to 29.57\% when looking at 20\% of the data, this is still better than BEER.
% \liz{an BEER? what?}

%Instead of measuring the similarity on the encoder hidden states, we can also measure the similarity on the decoder hidden states. Therefore, 
We performed the same experiment using the target hidden representations. Again, we investigated the influence of the bottleneck size and achieved the best performance with a bottleneck size of $256$ (Dec Auto 256).  Reasonably, the target hidden states contain more information about the sequence-to-be-generated than the source states. 

Finally, when combining the output probability with the decoder hidden states (Dec Auto 256 + Prob), we are able to achieve the best performance. Again, it is better to use the autoencoder on the decoder hidden state than on the encoder hidden state. %This indicates that in this task, the distance to seen training examples is important.
It is worth noting that the pseudo-labels perform very well when including the posterior probabilities. Interestingly, we see a clear drop in performance between oracle and pseudo-labels when not using the posterior probabilities.

\begin{table*}[ht]
 \begin{center}
   \begin{tabular}{|l|l|c|c|c|c|} \hline 
Model & Alignment & 10\% & 20\% & Oracle & Pseudo-label \\ \hline
Prob & Giza DE-EN &  25.15 & 34.09 & 19.33 & 17.73 \\
Prob & Giza EN-DE &  28.90 & 39.86 & 22.58 & 21.18\\ \hline
Enc Auto 256 & & 28.78 & 49.58 & 23.54 & 16.31\\ \hline
Dec Auto 256 & Giza EN-DE & 32.21 & 49.51 & 24.85 & 19.49\\
Enc Auto 256 + Prob & Giza EN-DE & 29.75 & 40.44 & 22.97 & 22.37\\ 
Dec Auto 256 + Prob & Giza EN-DE & 32.21 & 47.38 & 24.54 & 24.37\\ \hline
Prob & 256 & 32.34 & 44.59 & 25.16 & 24.25\\
%Prob & 256 TS & 30.27 & 41.15 & 23.64 & \\
Prob & 512 & 32.79 & 44.91 & 25.45 & 24.39\\
%Prob & 512 TS & 30.65 & 41.87 & 23.96 & \\
Prob & 2048 & 32.27 & 44.91 & 24.98 & 23.85\\
%Prob & 2048 TS & 30.53 & 42.06 & 23.81 & \\
Prob & 8192 &  33.38 & 46.34 & 25.63 & 24.33\\ \hline
%Prob & 8192 TS &  30.65 & 41.87 & 23.77 & \\
%Dec Auto 256 MSE & Prob 256 ST & 33.18 & 51.00 & 25.27 \\
%Dec Auto 256 MSE & Prob 512 ST & 33.12 & 51.39 & 25.40 &\\
%Dec Auto 256 MSE & Prob 2048 ST & 34.35 & 52.37 & 26.29 &\\
Dec Auto 256 & 8192 & 33.89 & 53.27 & 26.23 & 21.65\\ \hline
%Dec Auto 256 Prob & Prob 256 ST &  34.67 & 50.36 & 26.77 &\\
%Dec Auto 256 Prob & Prob 512 ST &  35.06 & 51.20 & 26.96 &\\
%Dec Auto 256 Prob & Prob 2048 ST & 35.00 & 51.46 & 26.83 &\\
Dec Auto 256 + Prob & 8192 & \textbf{35.58} & \textbf{52.62} & \textbf{27.21} & \textbf{27.02}\\ \hline

\end{tabular}   
\end{center}
\caption{\label{table:smallMT:source} Source word confidence for MT. First two columns: Percentage of found errors when selecting 10\% and 20\% of the data; Final two columns: F-score when using oracle threshold and thresholds optimized on pseudo-labels}
\end{table*}

\begin{table*}[ht]
 \begin{center}
   \begin{tabular}{|ll|cc|cc|cc|cc|} \hline 
\multicolumn{2}{|c|}{Generation} & \multicolumn{2}{|c|}{4 layer} & \multicolumn{2}{|c|}{8 layer} & \multicolumn{2}{|c|}{12 layer} & \multicolumn{2}{|c|}{24 layer}\\ \hline
&& SWB & CH & SWB & CH & SWB & CH & SWB & CH \\
\multicolumn{2}{|c|}{WER} & 20.8 & 33.2 & 14.8 & 25.5 & 13.0 & 23.9 & 12.1 & 23.0 \\ \hline \hline
Layer & Methods & 10\% & 20\% & 10\% & 20\% & 10\% & 20\% & 10\% & 20\%\\ \hline
4 & prob & 23.29 & 43.63 & 26.46 & 46.89 & 26.60 & 48.12 & 27.63 & 48.96 \\
4 & auto+prob & 23.50 & 43.87 & 26.45 & 46.96 & 26.67 & 48.14 & 27.64 & 49.00 \\
8 & prob & 30.00 & 53.08 & 29.37 & 51.83 & 32.65 & 54.76 & 32.46 & 55.02 \\
8 & auto+prob & 30.00 & 53.20 & 29.67 & 53.80 & 32.63 & 57.10 & 32.43 & 57.87 \\
12 & prob & 30.09 & 54.80 & 34.59 & 56.44 & 31.13 & 54.71 & 34.67 & 56.61\\
12 & auto+prob & 30.97 & 54.87 & 34.62 & 58.61 & 31.37 & 57.09 & 34.70 & 59.31 \\
24 & prob & 31.21 & 55.40 & 35.34 & 57.29 & 35.91 & 57.54 & 31.95 & 55.30 \\
24 & auto+prob & 31.65 & 57.26 & 36.05 & 61.15 & 36.73 & 60.97 & 37.42 & 60.77 \\ \hline

\end{tabular}   
\end{center}
\caption{\label{table:SW} Confidence estimation on ASR using different ASR systems for output predictions and confidence estimation: Found errors when selection 10\% and 20\% of the data}
\end{table*}

Moreover, we evaluated methods to identify source words with low confidence. The results for these experiments are summarized in Table \ref{table:smallMT:source}. %This is especially interesting for use cases where a system user only knows the source language. By marking the problematic source words, the user is able to reformulate the source sentence to achieve a better translation performance.
In this case the baseline is to map the posterior probabilities to the source sentence using a GIZA \cite{Och03} alignment. Again, we evaluate the approach with the same four scores.  As shown in the first two lines, the Giza alignment from source to target performance clearly better than the one from target to source. Therefore, in the remaining experiments, we only evaluate approaches using the source to target alignment.

By using the training-test distance approximated by the autoencoder on the encoder states (Enc Auto 256), we directly have an estimate on the source side and so do not need to map target estimates to the source side. In this case, we see improvements over using the posterior probabilities. Again, the pseudo labels perform not as well without using the posterior probabilities.
Next, we map the other three measures, decoder hidden states and the combination of encoder or decoder states and output probabilities, using the Giza alignment to the source. Interestingly, this time, solely using the approximation of the training-test similarity is even better than the combination with the output probabilities. The best system is achieved by  the autoencoder of the decoder states (Dec Auto 256). We see improvements by 3\% and 10\% over the posterior probabilities when looking at 10\% and 20\% of the data.
Finally, we tried to use an internal alignment instead of the Giza alignment. Therefore, we predict the decoder hidden states based on the encoder hidden states as described in Section \ref{alignment}. Again, we investigated different sizes for the hidden states used to map the posterior probabilities. As shown in Table \ref{table:smallMT:source}, all the models perform better than the GIZA alignment. We can further improve the quality by using a larger hidden layer. Since we need to learn a very complex mapping from source to target hidden states, a larger layer is better. The best performance is achieved using a layer of 8192 hidden units.

In the end, we also used the same model to map the autoencoder predictions. The combination of all three methods leads to the best results (Dec Auto 256 + Prob, 8192). By looking only at 10\% of the words, we are able to find more than 35\% of the errors and for 20\% of the words we identify more than half of the errors.

\subsection{Speech recognition results}

 The ASR results for this task are summarized in Table \ref{table:SW}. We present the percentage of found errors when looking at 10 and 20 percent of the data. In each column, we estimate the quality of one output generated by the different models. Each row represents the results when using one model to estimate the quality of the different outputs.

Here we evaluated four different models with increasing transcription quality. The only difference between the models are the number of hidden layers. We investigated models using $4, 8, 12$ and $24$ layers. In this task, the test set consists of two subsets. The best model achieves a word error rate of $12.1$ and $23.0$ on the two subsets, respectively.

Again, we use the output probabilities as well as the combination of the autoencoder and the output probabilities. %Since in the previous MT experiments the best results were achieved with a bottleneck size of half the input size,
We again use half the input size for the bottleneck size.
Firstly, as shown in the MT experiments, we can improve the quality estimation by combining the posterior probabilities and the autoencoder approach. In all configurations, the combination performs better or similar than the posterior probability.

Secondly, the better models are able to better estimate the confidence on the same output. In most cases, the performance can be improved by using a more complex model to estimate the confidence. One exception is the estimation of its own output. % In this case, often the performance drops.

Finally, the estimation of the distance between training and test data mainly helps when using stronger models, both for the generation of the output and for confidence estimation. 
Furthermore, this method also removes the effect of models performing worse than their own output. 
% \liz{mm I don't understand what you meant by this sentence}

\section{Related Work}

Prior work has investigated confidence measurement for speech recognition models~\cite{Siu1999}, and statistical machine translation models using either word-level posteriors~\cite{ueffing2007} or external models~\cite{Gandrabur2003}. Deep learning models have also received attention on uncertainty and confidence measurement recently: \cite{Gal2016Theoretically} formulate neural network models with dropout as Bayesian models to obtain uncertainty based on sampling methods. Specifically, for neural machine translation models or other sequence-to-sequence models, quality estimation has remained as a topic of concern. While most prior research focused on developing confidence measures for a general system using external features \cite{specia2019}, this works concentrates on estimating the confidence of a specific system by making use of the information available in the internal representation of the network.

\section{Conclusion}

In this work, we investigated the ability of sequence-to-sequence models to model their confidence in their decisions. We performed experiments  using these models for two tasks: machine translation and speech recognition. 

We analyzed the influence of train-test mismatch on quality estimation. By approximating this mismatch using an autoencoder and combining it with the posterior probabilities, we are able to improve confidence estimation over a strong baseline. We showed that it is better to measure the mismatch on the decoder hidden states than on the encoder hidden states.

Secondly, we also investigated methods to predict how well each individual source token is translated by a given model. 
In this case, measuring the train-test mismatch was even more important. Furthermore, we present an approach to infer the internal alignment of complex sequence-to-sequence models. Using this alignment instead of a state-of-the-art external alignment for mapping target confidence measure to source tokens clearly improved the quality of the confidence measure for source words

\section*{Acknowledgments}
\vspace{-0.5em}
The project ELITR leading to this publication has received funding from the European Union’s Horizon 2020 Research and Innovation Programme under grant agreement N\textsuperscript{\underline{o}} 825460.
We thank Elizabeth Salesky for the constructive comments.

\bibliography{acl2019}
\bibliographystyle{acl_natbib}

\end{document}